\documentclass[11pt,a4paper]{article}
\usepackage{times,latexsym}
\usepackage{url}
\usepackage[T1]{fontenc}

\usepackage[acceptedWithA]{tacl2021v1}

\usepackage{xspace,mfirstuc,tabulary}

\usepackage{graphicx}

\usepackage{algorithmic,algorithm}
\renewcommand{\algorithmiccomment}[1]{\bgroup\hfill$\triangleright$~#1\egroup}
\usepackage[linguistics]{forest} 
\usepackage{synttree}
\usepackage{subcaption}
\usepackage{tikz-dependency}
\usepackage[cjk]{kotex}

\usepackage{amsmath,amssymb}

\usepackage{arydshln}
\usepackage{booktabs}

\usepackage{amsthm}
\newtheorem{theorem}{Theorem}
\newtheorem{lemma}{Lemma}

\DeclareMathOperator*{\argmax}{argmax}

\newif\iftaclinstructions
\taclinstructionsfalse 
\iftaclinstructions
\renewcommand{\confidential}{}
\renewcommand{\anonsubtext}{(No author info supplied here, for consistency with
TACL-submission anonymization requirements)}
\newcommand{\instr}
\fi

\iftaclpubformat 

\else

\fi

\newcommand*{\jp}[1]{{\color{black}#1}}

\title{Reformulation and Equivalence of Bock's Algorithm for Exact Dependency Parsing}

\author{Yuxi Wang \\
UBC\\
Canada\\
{\tt johnyuxiwang@students.ubc.ca} 
\And 
Jungyeul Park\thanks{~~Corresponding author.}\\
KAIST\\
South Korea\\
{\tt jungyeul@kaist.ac.kr}
}
\date{}

\begin{document}
\maketitle

\begin{abstract}
Bock's algorithm provides a matrix-based alternative to Chu--Liu/Edmonds (CL/E) for exact dependency parsing. We present a reformulation that retains original node and arc identities through component labels, column adjustments, and star-and-bar transfers. In exact arithmetic, we prove operational equivalence to original Bock under matching scan and tie rules, and establish correspondence with corrected Tarjan under a prescribed component schedule. A separate analysis establishes optimal-value equivalence with Dozat-derived CL/E under the stated score and single-root conventions. 
We integrate reformulated Bock into a biaffine dependency parser and evaluate decoding and structured training on the Penn Treebank and Chinese Treebank. 
These results establish Bock as an alternative exact oracle for dependency parsing and structured training, while distinguishing equality of optimal values from identity of returned trees and training trajectories.
\end{abstract}

\section{Introduction}
\label{sec:introduction}

Exact arborescence inference is a standard component of graph-based dependency parsing. Given scores for individual head--dependent arcs, the decoder selects a highest-scoring tree subject to the required structural constraints \citep{mcdonald-etal-2005-non}. The Chu--Liu/Edmonds (CL/E) algorithm provides the classical basis for this computation \citep{chu-liu-1965-on-the-shortest,edmonds-1967-optimum}. Neural parsers retain this separation between learned arc scores and exact tree inference, including parsers using biaffine scoring \citep{dozat-manning-2017-deep}. CL/E resolves directed circuits among provisionally selected arcs through contraction and recovers original arcs through expansion. 
Recent comparisons of implementations of Edmonds' algorithm show that practical performance depends on graph structure and the data structures used \citep{bother-kibig-weyand-2023-efficiently}.

Frederick C. Bock's 1971 algorithm \citep{bock-1971-an-algorithm} solves the rooted arborescence problem through a matrix-based representation. It retains the original matrix and node indices, represents contracted components by shared labels, and selects entering arcs using column adjustments. Starred predecessors record provisional links, while barred references guide the transfers that replace them. These operations are interleaved in the original Algol procedure through label-directed control flow. Their interaction requires a precise account to establish that a modern implementation preserves the original computation.\footnote{Both Chu--Liu and Bock minimize the cost of a rooted arborescence \citep{chu-liu-1965-on-the-shortest,bock-1971-an-algorithm}. Edmonds develops a maximum-weight branching algorithm and gives reductions for both minimum- and maximum-weight rooted arborescences \citep{edmonds-1967-optimum}. The CL/E parser used here maximizes tree scores. Negating arc scores aligns these objectives when the admissible arcs and root constraints agree.}

We give an executable reformulation that separates candidate search, column adjustment, backward tracing, contraction, and star-and-bar transfer while preserving their dependencies and assignment order. In exact arithmetic, we prove its operational correspondence with Bock's original procedure under matching input conventions, scan order, and tie resolution. Corresponding maintained states yield the same candidate choices, component partitions, and final original arcs.

We also establish the relationship between the reformulation and CL/E. Under a prescribed component schedule and a common ordering of original arcs for resolving ties, we prove correspondence with Tarjan's implementation of Edmonds' method \citep{tarjan-1977-finding}, using the corrected value and recovery rules \citep{camerini-fratta-maffioli-1979-a-note}. This correspondence aligns component formation, adjusted selection values, and recovered arc sets, including when the optimum is nonunique. A separate analysis establishes optimal-value equivalence with the Dozat-derived CL/E implementation used in Stanza \citep{qi-etal-2020-stanza} under the stated score conventions and exactly-one-root-child constraint. These results distinguish correspondence between coupled executions from equality of optimal values and identity of native returned trees.

We integrate reformulated Bock into Stanza for prediction and loss-augmented inference. Experiments on dependency conversions of the Penn Treebank (PTB) \citep{marcus-etal-1993-building,de-marneffe-etal-2006-generating,schuster-manning-2016-enhanced} and Chinese Treebank (CTB) \citep{xue-etal-2005-ctb,xue-2007-tapping} compare local cross-entropy with CL/E- and Bock-based structured-hinge training under shared architecture, features, and optimization settings. Each condition uses its corresponding decoder for development selection and final prediction. 
On identical scores from the selected PTB and CTB local-CE models, CL/E, original Bock, and reformulated Bock return identical development and test parses and objective values. These results establish Bock as an alternative exact oracle for dependency parsing and structured training.

\section{Exact inference in dependency parsing}
\label{sec:exact-inference}

For a sentence of $n\geq1$ tokens, index the tokens by $1,\ldots,n$ and let node $0$ denote an artificial \textsc{root}. An arc $i\to j$ assigns node $i$ as the head of token $j$. {An arc-factored parser assigns a finite real score $S[i,j]$ to each admissible arc $i\to j$ and scores a tree by summing its arc scores \citep{mcdonald-etal-2005-non}. Rows of $S$ index heads and columns index dependents.} The scores may depend on the entire sentence; arc factorization concerns their additive contribution to the tree objective.

A rooted spanning arborescence contains exactly one incoming arc for each token, no incoming arc for node $0$, and a directed path from node $0$ to every token. These conditions allow several tokens to attach directly to the artificial root. A single-root dependency tree additionally requires exactly one outgoing arc from node $0$: precisely one token has head $0$. This additional constraint must therefore be enforced when using an arborescence algorithm for single-root dependency parsing. No projectivity constraint is imposed.

Self-loops and arcs entering node $0$ are forbidden. Any further restrictions are imposed identically for both decoders. Forbidden arcs are excluded from selection and represented by $-\infty$ in the score matrix; {the column corresponding to node $0$ is unused in the parsing objective.} Let $\mathcal{T}_1$ denote the set of single-root dependency trees containing only admissible arcs. Exact decoding solves
\begin{equation}
  \widehat{T}\in\argmax_{T\in\mathcal{T}_1}
  \sum_{(i,j)\in T} {S[i,j]}.
  \label{eq:parsing-objective}
\end{equation}
If $\mathcal{T}_1$ is empty, the instance is infeasible.

{Bock's procedure minimizes costs using the same matrix orientation: rows index initial nodes (heads), and columns index terminal nodes (dependents).} For each admissible arc, we therefore set
\begin{equation}
  C[i,j]=-{S[i,j]},
  \label{eq:score-cost-conversion}
\end{equation}
and assign $+\infty$ to forbidden arcs. Throughout, $+\infty$ and $-\infty$ are symbolic exclusion sentinels: forbidden arcs are omitted from candidate sets before cost or score arithmetic. For every feasible tree, its total cost is the negative of its total score. Consequently, minimum-cost and maximum-score decoding have exactly the same optimal trees when their admissible arcs and root constraints agree.

Equality of optimal values, identity of returned trees, and correspondence between executions are distinct claims. On a feasible instance, two exact decoders for Equation~\eqref{eq:parsing-objective} attain the same optimal score in exact arithmetic. They necessarily return the same tree when the optimum is unique, but may select different trees when several optima exist. Correspondence between executions requires a relation between their maintained states and a proof that their operations preserve that relation. Neither equal objective values nor identical final trees alone establishes such a correspondence.

\section{Reformulating Bock's algorithm}
\label{sec:bock}

\subsection{Original procedure and maintained state}

Bock's procedure computes a minimum-cost arborescence with a designated root \citep{bock-1971-an-algorithm}. We retain the nodes $0,\ldots,n$ and cost convention of Section~\ref{sec:exact-inference}, with finite costs for admissible arcs and $+\infty$ for forbidden arcs. {The entry $C[i,j]$ is the cost of assigning head $i$ to dependent $j$.} The root is node $0$; restricting it to exactly one child requires a separate constraint. Appendix~\ref{sec:bock-original} gives the original procedure. {For the comparison, take its origin to be node $1$ and subtract one from every node index. Represent the original empty predecessor value $0$ and empty bar pair $(0,0)$ by $\bot$, distinguishing them from node $0$.}

The procedure processes nonroot columns in increasing order. The array $U_1[j]$ records the cumulative adjustment applied to column $j$, and $I^*[j]=i$ records a provisional, or \emph{starred}, arc $i\to j$. A barred pair {$B[j]=(\bar I[j],\bar J[j])$} stores an original arc reference. {When $B[j]\neq\bot$, $\bar I[j]$ and $\bar J[j]$ are respectively the head and dependent of the referenced arc; $\bar J[j]$ need not equal $j$.} Backward tracing writes the current candidate into empty bar entries, and transfers may subsequently redirect these references. If installing an arc displaces an existing star, the bar read at its {dependent} supplies the next candidate. In the original procedure, equal positive $\mathrm{SPAN}$ labels represent a contracted component, and negative labels temporarily mark components visited during a backward trace.
\jp{Algorithm~\ref{alg:bock} presents our reformulation of Bock's procedure.}

\begin{algorithm}[!ht]
\footnotesize
\caption{Reformulated Bock procedure}
\label{alg:bock}
\begin{algorithmic}[1]
\REQUIRE Cost matrix $C$ on nodes $0,\ldots,n$
\ENSURE Minimum-cost predecessor array, or \textsc{Infeasible}
\STATE For all $j$, set $U_1[j]\gets0$, $I^*[j]\gets\bot$, $B[j]\gets\bot$, and $\mathrm{comp}[j]\gets j$.
\FOR{$K=1,\ldots,n$}
  \REPEAT
    \STATE $H\gets\mathrm{comp}[K]$.
    \STATE Choose the first minimum $(i,j)$ of $C[i,j]-U_1[j]$ satisfying \eqref{eq:bock-candidates}.
    \STATE If no candidate exists, \textbf{return} \textsc{Infeasible}.
    \STATE $DU\gets C[i,j]-U_1[j]$.
    \STATE Add $DU$ to $U_1[q]$ for every $q\leq K$ with $\mathrm{comp}[q]=H$.
    \STATE $v\gets i$; $\mathrm{TRACE}\gets\varnothing$.
    \WHILE{$I^*[v]\neq\bot$ and $\mathrm{comp}[v]\neq H$}
      \STATE Add every $q\leq K$ with $\mathrm{comp}[q]=\mathrm{comp}[v]$ to $\mathrm{TRACE}$.
      \STATE If $B[v]=\bot$, set $B[v]\gets(i,j)$.
      \STATE $v\gets I^*[v]$.
    \ENDWHILE
    \STATE If $\mathrm{comp}[v]=H$, set $\mathrm{comp}[q]\gets H$ for every $q\in\mathrm{TRACE}$.
  \UNTIL{$\mathrm{comp}[v]\neq H$}
  \STATE $e\gets(i,j)$; $d\gets\bot$.
  \REPEAT
    \STATE $(i,j)\gets e$.
    \STATE For every $q\leq K$ with $B[q]=e$, set $B[q]\gets d$.
    \STATE $f\gets B[j]$; $B[j]\gets d$.
    \STATE $a\gets I^*[j]$; $I^*[j]\gets i$.
    \STATE If $a\neq\bot$, set $d\gets(a,j)$ and $e\gets f$.
  \UNTIL{$a=\bot$}
\ENDFOR
\STATE \textbf{return} $I^*$.
\end{algorithmic}
\end{algorithm}

\subsection{Reformulated procedure}

We retain $U_1$, $I^*$ and $B$, and replace $\mathrm{SPAN}$ by an array $\mathrm{comp}$ and an explicit trace set. Component labels remain unchanged during tracing. A contraction assigns the active component's label to the traced components, preserving the represented partition without creating a fresh label.

For the current column $K$, {let $H=\mathrm{comp}[K]$ be the label of its active component.} Candidate selection considers precisely the arcs satisfying
\begin{equation}
\begin{gathered}
  1\leq j\leq K,\quad \mathrm{comp}[j]=H,\\
  0\leq i\leq n,\quad
  \mathrm{comp}[i]\neq H,\quad C[i,j]<+\infty.
\end{gathered}
\label{eq:bock-candidates}
\end{equation}
{Among these arcs, the procedure minimizes $C[i,j]-U_1[j]$, examining dependents in increasing $j$ and, for each dependent, heads in increasing $i$, and retaining the first minimum.} Adding this minimum to the adjustments of the active component makes the selected candidate's adjusted cost zero and leaves every current candidate with nonnegative adjusted cost.

The backward trace follows starred predecessors from the candidate's {head $i$}. Returning to the active component identifies a circuit and triggers contraction. Reaching an empty predecessor instead initiates a transfer. Each transfer first redirects every processed bar referring to the current candidate, then reads the bar at its {dependent $j$}, replaces that bar, and installs the candidate. The global replacement must precede the local read, since it may change the reference that determines the next candidate.

\paragraph{Example}
\jp{Consider nodes $0,1,2$, with root $0$ and admissible arcs $0\to1$, $0\to2$, $1\to2$, and $2\to1$ of costs $50$, $60$, $10$, and $20$, respectively. This is the instance in Appendix~\ref{sec:bock-original} with every node index reduced by one. At $K=1$, the procedure installs $2\to1$ with $U_1[1]=20$. At $K=2$, selecting $1\to2$ gives $U_1=(0,20,10)$. The backward trace records $\mathrm{TRACE}=\{1\}$ and $B[1]=(1,2)$, then returns to the active component, detecting the circuit $1\to2\to1$. Contraction yields $\mathrm{comp}=(0,2,2)$.}
\jp{The renewed search compares adjusted costs $50-20=30$ for $0\to1$ and $60-10=50$ for $0\to2$. It selects $0\to1$ and updates $U_1$ to $(0,50,40)$. The transfer reads and clears $B[1]$, replaces the star $2\to1$ by $0\to1$, then installs the stored candidate $1\to2$ and sets $B[2]=(2,1)$. The final array $I^*=(\bot,0,1)$ defines the arborescence $0\to1\to2$ of cost $50+10=60$; the residual bar entry does not affect the selected tree.}
\jp{The selections and transfer match the original procedure. The reformulation records the backward trace explicitly in $\mathrm{TRACE}$ rather than marking component labels negatively, and stores each bar pair in a single entry of $B$ rather than in parallel arrays.}

\subsection{Faithfulness to original Bock}

{The reformulation preserves the operations of the original Bock procedure while changing how components and backward traces are represented. The original $\mathrm{SPAN}$ serves two purposes: equal absolute labels identify members of the same component, while negative labels temporarily mark components visited during a backward trace. The reformulation separates these roles. The array $\mathrm{comp}$ maintains component membership without changing labels during tracing, and $\mathrm{TRACE}$ collects all processed vertices belonging to each visited component. Thus, negating a component's $\mathrm{SPAN}$ label in the original procedure corresponds to adding its members to $\mathrm{TRACE}$. If the trace returns to the active component, the original procedure assigns a fresh label to the active and negatively marked components, whereas the reformulation assigns the active label to all vertices in $\mathrm{TRACE}$. Both merge exactly the same components. If the trace terminates without a circuit, restoring the negative $\mathrm{SPAN}$ labels corresponds to simply discarding $\mathrm{TRACE}$, leaving the component partition unchanged.}

{Assume exact arithmetic and the same scan order and first-minimum tie rule. Initially, the column adjustments, starred predecessors, and barred pairs agree after the index conversion, and the component arrays represent the same partition. Whenever these relations hold before a candidate search, the procedures examine the same ordered candidate set with identical adjusted costs, select the same minimum, and update $U_1$ identically.}
{The backward traces follow the same starred predecessors and fill the same empty bar entries. The transfer phase likewise preserves the original order of global bar replacement, local bar read, and star update, producing identical $B$ and $I^*$. 
{Induction over candidate searches, backward-trace steps, transfer iterations, and column advances therefore establishes these relations at corresponding phase boundaries. The procedures return the same predecessor array and objective value, or both report infeasibility.}
}

\subsection{Computational complexity}
\label{sec:bock-complexity}

Assume unit-cost arithmetic, comparisons, and array access, with $\mathrm{TRACE}$ represented by a Boolean mask. Each successful candidate search either completes a column or contracts at least two components. There are $n$ columns and at most $n-1$ contractions, so the procedure performs $O(n)$ searches; an empty candidate set terminates the computation. Each candidate scan costs $O(n^2)$. \jp{A backward trace follows a path in the starred forest and therefore visits at most $n+1$ vertices.} Scanning the component labels at each visited vertex therefore costs at most $O(n^2)$ per trace. Column adjustments and component relabelling each cost $O(n)$.

\jp{The transfer completing a column uses at most $n$ iterations. Within a contracted component, it first acts in the constituent containing the entering arc's dependent and, if different, then in the constituent containing the component's previous internal root, the vertex without an internal starred predecessor. The same decomposition applies recursively within each constituent. Because the two constituents are disjoint, induction over nested contractions shows that each dependent receives at most one new star during the transfer.} Each iteration scans at most $n$ bar entries and performs a constant number of local assignments, giving $O(n^2)$ time per transfer. Consequently, Algorithm~\ref{alg:bock} runs in $O(n^3)$ time. The cost matrix occupies $O(n^2)$ storage, while the maintained arrays and trace mask require $O(n)$ additional storage. \jp{Bock's original procedure has the same $O(n^3)$ time and $O(n^2)$ total-space upper bounds: its candidate scans and transfers have the same asymptotic costs, while span marking and relabelling require at most $O(n^2)$ work per trace.} The same upper bounds hold on an infeasible instance, since the procedure terminates at its first empty candidate set.

\section{Equivalence with CL/E}
\label{sec:equivalence}

\subsection{Correspondence with corrected Tarjan}

The comparison with Tarjan concerns the rooted arborescence problem before imposing the single-root constraint. We assume that an arborescence rooted at node $0$ exists. Tarjan's original formulation maximizes total arc weight over branchings, which may contain several trees \citep{tarjan-1977-finding}. We use its sign-reversed, minimum-value formulation and enforce maximum cardinality before minimizing cost. Initialize it with
\[
  d(i,j)=C[i,j]-\Lambda
\]
for every admissible arc. Here $\Lambda$ is a positive formal symbol enforcing cardinality priority on admissible arcs: affine expressions are ordered first by their coefficient of $\Lambda$ and then by their real part. {It is distinct from $+\infty$, which indicates excluded arcs.} A branching has at most $n$ arcs, and a feasible rooted arborescence attains this cardinality. Since no arc enters node $0$, every branching with $n$ arcs is rooted at $0$. Minimizing the transformed objective therefore selects precisely the minimum-cost rooted arborescences.

We use Tarjan's corrected value comparisons and recovery procedure \citep{camerini-fratta-maffioli-1979-a-note}. A root component is a strongly connected component of the selected graph with no selected arc entering it from outside. Tarjan extracts a minimum-current-value unexamined arc whose {dependent} lies in a chosen root component. An internal arc is discarded; an external arc is selected. If the selection closes a circuit of components $Q_1,\ldots,Q_m$, with circuit arc $e_\ell$ entering $Q_\ell$, the update is
\begin{equation}
\begin{aligned}
  h&=\max_{\ell}d(e_\ell),\\
  d(i,j)&\gets d(i,j)-d(e_\ell)+h,
  \qquad j\in Q_\ell.
\end{aligned}
\label{eq:tarjan-update}
\end{equation}
The update applies only to unexamined arcs. Each selected arc retains its value at extraction.

Bock processes columns in increasing order. We prescribe Tarjan's component schedule to follow Bock's active component, continuing with the merged component after a contraction. Pair each Bock candidate search with Tarjan's state immediately before its next external extraction, after any internal extractions. Both executions resolve equal-value choices by increasing original dependent index and then increasing original head index. All arc identities are retained through contraction.

For each contraction, consider its immediate constituent components $Q_1,\ldots,Q_m$ and its original circuit arcs $e_1,\ldots,e_m$, including the closing arc. These records describe the execution history without changing either algorithm. {For an original arc $e=(i,j)$, write $\operatorname{dep}(e)=j$ for its dependent.} Define the internal expansion rooted at $x$ by $E(\{x\},x)=\varnothing$ and, when $x\in Q_s$, by
\begin{equation}
\begin{aligned}
  E(Q,x)
  &=E(Q_s,x)\\
  &\quad{}\cup\bigcup_{\ell\neq s}
  \left(
    \{e_\ell\}\cup
    E(Q_\ell,{\operatorname{dep}(e_\ell)})
  \right).
\end{aligned}
\label{eq:bock-recursive-expansion}
\end{equation}
Removing $e_s$ from the constituent circuit leaves an out-tree rooted at $Q_s$. Induction on the contraction history therefore shows that $E(Q,x)$ is an internal arborescence rooted at $x$.

The corresponding recursive bar representation is defined as follows. If the internal root of $Q$ lies in $Q_s$, each constituent $Q_\ell$ with $\ell\neq s$ has internal root {$\operatorname{dep}(e_\ell)$}, external star $e_\ell$, and root bar $e_s$. The constituent $Q_s$ inherits the external star and root bar of $Q$. The same description holds recursively inside every constituent. Its internal stars are exactly $E(Q,x)$, where $x$ is its internal root. Stars and bars in this representation are identified by their original arcs.

\begin{lemma}[Transfer and recursive expansion]
\label{lem:bock-transfer-expansion}
Suppose a recorded component $Q\subseteq\{1,\ldots,K\}$ has the recursive bar representation, with external star $p$ and root bar $b$. Assume that no bar outside $Q$ names an internal circuit arc in its history. Here $p$ is empty or enters $Q$, and $b$ is empty or has its {dependent} outside $Q$.

Let the current candidate $e$ enter $Q$ at $x$, and let the preceding displaced arc $z$ be empty or have its {dependent} outside $Q$. A finite consecutive segment of the transfer in Algorithm~\ref{alg:bock} installs $e$ at $x$, replaces the internal stars by $E(Q,x)$, and installs root bar $z$. It returns the old external star $p$ as the displacement and reads the old root bar $b$ as the next candidate. Outside $Q$, only the initial global substitution $e\mapsto z$ can change bars, and no star changes. The complete transfer stops at this segment precisely when $p$ is empty.
\end{lemma}

\begin{proof}
An original arc belongs to the circuit of at most one contraction: after its endpoints are merged, it is internal to every later component containing them. The recursive representation and the locality hypothesis therefore identify references to an omitted circuit arc precisely at the nonroot constituent roots of that contraction. The same locality holds inside each constituent, since {the arcs referenced by nonempty inherited outer bars have their dependents outside that constituent.}

Induct on the contraction history of $Q$. {For a singleton, $b\neq e$: this is immediate if $b$ is empty; otherwise, their dependents lie on opposite sides of $Q$.} One transfer iteration reads $b$, installs $z$ and $e$, and returns $p$, as required.

Otherwise, let $Q_s$ contain the old internal root and let $x\in Q_t$. Apply the induction hypothesis inside $Q_t$. Since $e$ enters $Q$ from outside, its initial substitution changes no bar in another constituent of $Q$. If $t=s$, the segment returns $p,b$, and the omitted circuit arc remains $e_s$.

If $t\neq s$, the first segment returns displaced arc $e_t$ and next candidate $e_s$. The displacement is nonempty, so the transfer continues inside $Q_s$, now with candidate $e_s$ and preceding displacement $e_t$. Its initial substitution $e_s\mapsto e_t$ redirects all remaining references at the other constituent roots. The induction hypothesis for $Q_s$ returns $p,b$. The new root constituent is $Q_t$, the omitted arc is $e_t$, and every other constituent has entering star $e_\ell$ and root bar $e_t$. This is precisely the recursive representation rooted at $x$.

Every substituted label after the initial substitution is internal to $Q$, so locality gives the asserted outside effect. Recursive segments act in smaller constituents, and the two constituents used when $t\neq s$ are disjoint. The segment is finite, using at most $|Q|$ iterations. Every intermediate nonroot constituent returns its nonempty circuit arc, so the stopping rule suppresses no required continuation.
\end{proof}

\begin{theorem}[Coupled Bock--Tarjan executions]
\label{thm:bock-tarjan}
In exact arithmetic, the prescribed schedule is well defined. At every paired search, the component partitions coincide, the active component is a root component, and both external candidate sets contain exactly the admissible original arcs entering it from outside. These arcs are unexamined by Tarjan. The executions select the same external arc and contract the same directed component circuit, with the same original circuit arcs.

For every current component $Q$, an offset $\beta_Q$ satisfies
\begin{equation}
  d(i,j)=C[i,j]-U_1[j]+\beta_Q
  \label{eq:component-offset}
\end{equation}
on every unexamined external arc entering $Q$.

After each completed prefix $K=0,\ldots,n$, corrected Tarjan recovery on the corresponding prefix record returns exactly Bock's starred arc set. The recovered branching has roots $\{0\}\cup\{K+1,\ldots,n\}$. In particular, the final rooted arborescences coincide even when the optimum is nonunique.
\end{theorem}

\begin{proof}
We use a joint induction over candidate searches and completed transfers. Maintain equality between Bock's components and Tarjan's strong components. Their common quotient is an out-forest, and Bock's crossing stars are exactly its selected arcs. Every current component has the recursive bar representation with an empty root bar. Its internal root carries the unique external entering star when the component is nonroot, and an empty star otherwise. During column $K$, the active component has internal root $K$, whose star is empty. The origin and unprocessed vertices remain singleton components with empty stars and bars. In addition to \eqref{eq:component-offset}, maintain that each nonroot component's offset is the frozen value of its entering selected arc.

Initially, all components are singletons, all stars and bars are empty, and \eqref{eq:component-offset} holds with $\beta_{\{v\}}=-\Lambda$. Suppose the invariants hold at a candidate search with active component $Q$. Its empty external star makes it a root of the common quotient. Moreover, $Q\subseteq\{1,\ldots,K\}$: the origin or an unprocessed singleton stops a backward trace before it can be marked or contracted. Thus Bock's restriction to processed columns excludes no arc entering $Q$.

Every admissible arc entering $Q$ from outside is unexamined by Tarjan. A discarded internal arc remains internal under subsequent mergers, while a previously selected external arc entering $Q$ would contradict rootness. Feasibility of the full rooted instance guarantees an entering arc for every nonempty component excluding node $0$. Internal extractions change neither the selected graph nor the external candidates or their values. Furthermore, every finite queue value retains the form $-\Lambda$ plus a real number under \eqref{eq:tarjan-update}, so Tarjan's branching stopping rule does not reject an eligible external arc.

Equation~\eqref{eq:component-offset} differs from Bock's comparison by a component-wide constant. The common original-arc tie order therefore gives the same choice $e$. {For $e=(i,j)$, the values before the column adjustment satisfy $DU=C[i,j]-U_1[j]$, so \eqref{eq:component-offset} gives $d(e)=DU+\beta_Q^{-}$.} After Bock's column adjustment, set
\[
  \beta_Q^{+}=\beta_Q^{-}+DU=d(e).
\]
This preserves \eqref{eq:component-offset} for the remaining external arcs, whose Tarjan values have not changed.

Inside each component, the backward starred path reaches its internal root. At a nonroot component it then follows the unique crossing selected arc; at a root component it stops. Hence Bock's projected trace is exactly the predecessor path in the common quotient forest. Adding $e$ creates a circuit precisely when its {head} belongs to the quotient tree rooted at $Q$. Both executions then identify the same path and closing arc, and merge the same components.

Only the internal root of each traced nonroot component has an empty bar. The trace writes $e$ at exactly these roots and preserves their internal bars. Consequently, contraction creates the recursive representation of the union with closing arc $e$ omitted and internal root $K$. The selected graph has exactly this new strong component, and its quotient remains an out-forest.

For each circuit constituent $Q_\ell$, its offset after the column adjustment equals $d(e_\ell)$: this follows from the nonroot invariant for the old circuit arcs and the selection identity for the closing arc. An unexamined arc entering the union from outside also entered its constituent from outside. Therefore \eqref{eq:tarjan-update} gives
\[
\begin{aligned}
  d^{+}(i,j)
  &=d^{-}(i,j)-d(e_\ell)+h\\
  &=C[i,j]-U_1[j]+h.
\end{aligned}
\]
The merged offset is $h$, and the induction continues at the next candidate search.

If no circuit forms, the trace ends at a different quotient root. It has written temporary references to $e$ only at the intervening nonroot component roots. Since $e$ is unexamined, it occurs in no earlier contraction record or bar. The initial transfer substitution $e\mapsto\bot$ clears exactly those temporary references outside $Q$. Apply Lemma~\ref{lem:bock-transfer-expansion} to $Q$ with empty external star, root bar, and preceding displacement. The transfer terminates inside $Q$, installs $e$, and restores its recursive representation rooted at {$\operatorname{dep}(e)$}. All outside stars are unchanged. Both quotient forests acquire the same crossing arc, and $Q$ becomes nonroot with offset $d(e)$. The next column is a singleton root, so all induction assertions are preserved.

It remains to identify corrected recovery. At a completed prefix $K$, every processed vertex has a star, while the origin and unprocessed vertices have none. The common quotient therefore has precisely the singleton roots $\{0\}\cup\{K+1,\ldots,n\}$. All selections have {dependents} at most $K$. On the graph containing only such arcs, these roots have no eligible incoming arcs, so the prefix construction is terminal.

In the corrected reconstruction forest, each selected original arc is a node \citep{camerini-fratta-maffioli-1979-a-note,bother-kibig-weyand-2023-efficiently}. An arc entering a singleton is its leaf; an arc entering a contracted component becomes the parent of all circuit nodes of the contraction that formed that component, including its closing node. Thus the subtree of an arc entering a component contains exactly the leaves for that component's original vertices.

Recovery retains a forest-root arc $e$ and deletes the path from the leaf for {$\operatorname{dep}(e)$} to that root. At each nested contraction, this path passes through the constituent containing that {dependent}. Its entering circuit arc is deleted, while the other constituent subtrees survive. Induction on the contraction history gives exactly
\[
  \{e\}\cup E(Q,{\operatorname{dep}(e)})
\]
for an arc entering $Q$. The surviving subtrees are disjoint, so their processing order does not affect the recovered arc set.

At a completed prefix, every current nonroot component has its entering selected arc as a reconstruction-forest root; all its internal contraction nodes lie below that root. The remaining root vertices have no incoming selections and require no root-circuit deletion. Recovery therefore gives precisely the recursive expansions represented by Bock's stars. This proves prefix equality, including the empty prefix. At $K=n$, only node $0$ remains a root, and the common branching is the claimed arborescence.
\end{proof}

\subsection{Equivalence with Dozat-derived CL/E}

The Stanza version used in our experiments implements CL/E by contraction and expansion, following Dozat's code \citep{qi-etal-2020-stanza}. We establish its optimal-value equivalence directly from that contraction rule and its single-root inference procedure.

At a contraction stage, let $p(j)$ be a maximum-scoring admissible head for dependent $j$, and let $Q$ be a directed cycle of these selections. {Put $\sigma_Q=\sum_{j\in Q}S[p(j),j]$.} Contracting $Q$ to node $q$ gives
\begin{equation}
\begin{aligned}
  {S'[i,q]}
    &=\sigma_Q+\max_{j\in Q}
      {\bigl(S[i,j]-S[p(j),j]\bigr)},\\
  {S'[q,k]}
    &=\max_{i\in Q}{S[i,k]},
\end{aligned}
\label{eq:stanza-contraction}
\end{equation}
{where $i$ in the first line is a head outside $Q$, and $k$ in the second is a dependent outside $Q$.} Other surviving scores remain unchanged, self-loops are excluded, and a maximum over no admissible arcs is $-\infty$. The maximizing original endpoints are retained for expansion.

An optimal rooted arborescence can be chosen to enter $Q$ exactly once. Indeed, take an optimal tree and an entering arc $u\to v$ whose root-to-$u$ path avoids $Q$. Such an arc exists by taking the first entry along a root path to $Q$. Retain this arc and replace the entering arc of every $j\in Q\setminus\{v\}$ by $p(j)\to j$. The root still reaches $u$ and hence $v$, and the selected cycle with $p(v)\to v$ removed reaches every vertex of $Q$. Every remaining vertex is reachable along an unchanged suffix of its former root path. The result is therefore an arborescence, and its score cannot decrease because each replacement uses a maximum-scoring incoming arc.

For a tree entering $Q$ once, the first line of \eqref{eq:stanza-contraction} accounts for its entering arc and retained cycle arcs. The second line accounts for each outgoing arc. Contracting such a tree gives a tree whose contracted score is at least its original score. Conversely, every contracted tree expands using the stored maximizing endpoints: its entering arc replaces exactly one cycle arc, all other cycle arcs are restored, and every outgoing arc receives its stored original head. The expanded tree has exactly the contracted score. Thus contraction preserves the optimum, and induction over contractions proves exact rooted-arborescence inference.

{Column-wise log-softmax subtracts a constant from each dependent column and therefore changes every tree score by the same total constant.} It preserves the maximizing trees and gives nonpositive admissible scores. If any dependent has at least two admissible heads, at least one resulting score is strictly negative.

\begin{theorem}[Single-root parsing equivalence]
\label{thm:single-root-equivalence}
Suppose $\mathcal T_1$ is nonempty and both decoders receive the same admissible arcs and scores. Assume that the finite scores are nonpositive and, when $n>1$, have a strictly negative minimum. In exact arithmetic, {the Dozat-derived CL/E and Bock single-root inference procedures} both solve Equation~\eqref{eq:parsing-objective} and attain the same optimal score.
\end{theorem}

\begin{proof}
For $n>1$, let $m<0$ be the minimum finite input score, with {the unused entry $S[0,0]$ set to zero}. {The CL/E procedure adds} $(n+1)m$ to every admissible root-arc score, following the root-reweighting approach \citep{stanojevic-cohen-2021-root}. Every rooted arborescence has $n$ arcs and an original score between $nm$ and $0$. The reweighted score of any single-root tree therefore exceeds that of any tree with $k\geq2$ root children by at least
\[
  nm-(k-1)(n+1)m\geq -m>0.
\]
Hence every reweighted optimum has exactly one root child. All single-root trees receive the same additive term, so their original ranking is preserved.

{The Bock single-root inference procedure} first solves the unrestricted rooted problem using \eqref{eq:score-cost-conversion}. If the returned optimum has one root child, it is already optimal over $\mathcal T_1$. Otherwise, the {procedure} enumerates admissible sole root children. For each child, it forces that root arc, excludes every other root arc, and solves the resulting rooted problem. These feasible subproblems partition $\mathcal T_1$ according to its unique root child. Selecting their best optimum therefore solves the single-root problem.

For $n=1$, the sole feasible root arc determines the unique tree in both procedures.
\end{proof}

\subsection{Complexity comparison}

\jp{For dense score matrices, the Dozat-derived CL/E procedure has an $O(n^3)$ time upper bound for rooted-arborescence inference: there are $O(n)$ contraction stages, each requiring at most $O(n^2)$ work, and expansion does not increase this bound. This matches the bound for Algorithm~\ref{alg:bock} proved in Section~\ref{sec:bock-complexity}.}

\jp{For single-root inference, CL/E reweights root arcs and performs one rooted-arborescence computation, retaining the $O(n^3)$ time bound \citep{stanojevic-cohen-2021-root}. The implemented Bock procedure solves one unrestricted rooted problem and, if the returned optimum has multiple root children, at most $n$ constrained subproblems, one per admissible sole root child. Its worst-case time is $O(n^4)$, with $O(n^2)$ peak storage when the subproblems are solved sequentially. If the initial optimum has one root child, the procedure returns it in $O(n^3)$ time. Under the assumptions of Theorem~\ref{thm:single-root-equivalence}, applying the same root reweighting to Bock's input scores would give $O(n^3)$ single-root inference; the experiments reported here use enumeration.}

\jp{These bounds concern the particular matrix procedures compared here. Tarjan's dense-graph implementation achieves $O(n^2)$ time for rooted-arborescence inference \citep{tarjan-1977-finding,bother-kibig-weyand-2023-efficiently}. The correspondence in Theorem~\ref{thm:bock-tarjan} identifies candidate choices, contractions, and recovered branchings, while their computational costs depend on the data structures and operations used to maintain them.}

\section{Experiments}
\label{sec:experiments}

\begin{table*}[!ht]
\centering
\footnotesize
\setlength{\tabcolsep}{4pt}
\begin{tabular}{lcccccc}
\toprule
& \multicolumn{3}{c}{PTB} & \multicolumn{3}{c}{CTB} \\
\cmidrule(lr){2-4}\cmidrule(lr){5-7}
Training & UAS & LAS & LCM & UAS & LAS & LCM \\
\midrule
Local CE
& $95.71 \pm 0.12$ & $94.79 \pm 0.10$ & $55.39 \pm 0.62$
& $90.07 \pm 0.44$ & $88.94 \pm 0.49$ & $39.72 \pm 0.94$ \\
Hinge (CL/E)
& $95.64 \pm 0.10$ & $94.71 \pm 0.10$ & $54.42 \pm 0.59$
& $90.12 \pm 0.24$ & $89.02 \pm 0.26$ & $39.23 \pm 1.08$ \\
Hinge (Bock)
& $95.72 \pm 0.05$ & $94.79 \pm 0.08$ & $54.55 \pm 0.85$
& $90.88 \pm 0.98$ & $89.80 \pm 1.08$ & $41.24 \pm 2.14$ \\
\bottomrule
\end{tabular}
\caption{Test scores (\%), reported as means $\pm$ sample standard deviations across three seeds. Local CE uses CL/E for development selection and final decoding; for all three PTB models, substituting original or reformulated Bock gives identical development and test predictions. Each hinge condition uses its corresponding decoder for loss-augmented training inference, development selection, and final decoding. Bock denotes the reformulated implementation.}\label{tab:attachment-results}
\end{table*}

We evaluate decoder substitution and structured training on the Penn Treebank (PTB) and Chinese Treebank (CTB), using Stanza's biaffine parser \citep{dozat-manning-2017-deep,qi-etal-2020-stanza}. All training conditions share the architecture, linguistic features, admissible arcs, and single-root constraint, with full BERT fine-tuning. Experimental settings, verification procedures, and supplementary comparisons appear in Appendix~\ref{sec:experimental-details}. Table~\ref{tab:attachment-results} reports test UAS, LAS, and labeled complete match (LCM). LCM requires every scored token in a sentence to have the correct head and dependency label.

\paragraph{Decoder substitution on identical scores}

For the three PTB local-CE models, one per seed, we apply CL/E, original Bock, and reformulated Bock to identical development and test score matrices. The admissible arcs, single-root constraint, and tie-breaking rules are shared across all three implementations. All three return identical heads, full relation labels, and original tree-score values on every evaluated sentence, yielding identical UAS, LAS, and complete-match scores for each model.

\paragraph{Structured training}

We also implement structured arc hinge loss using either CL/E or reformulated Bock. Relation-label cross-entropy and auxiliary losses are common to all conditions. Each hinge condition uses its corresponding decoder for loss-augmented training inference, development evaluation, and final prediction, so the comparison includes both the training oracle and development-dependent decisions.
On identical scores, gold trees, task costs, and admissible tree families, exact CL/E and Bock oracles define the same hinge objective. A unique loss-augmented optimum determines the same score subgradient; tied optima permit different valid subgradients. Identical training trajectories additionally require matched initialization, minibatches, stochastic and numerical computations, selected subgradients, and training decisions. Appendix~\ref{sec:structured-training} gives the precise conditions and proof.
Additional PTB experiments retain local cross-entropy while using CL/E, original Bock, or reformulated Bock for development evaluation and final prediction. In every seed, training losses differ before the first development evaluation, demonstrating variation without a structured-training oracle. Losses also differ before the first development evaluation in CL/E- and Bock-based hinge training on both PTB and CTB. Development decoding therefore cannot explain the initial variation, whose cause remains undetermined. The final accuracy differences do not isolate the effect of replacing the training oracle.

\paragraph{Parsing accuracy}

On PTB, hinge training with reformulated Bock exceeds CL/E hinge by 0.08 percentage points in both mean UAS and LAS, and matches local CE's mean LAS at the reported precision. Despite this similar attachment accuracy, mean LCM is 0.84 percentage points below local CE for Bock hinge and 0.98 points below local CE for CL/E hinge.
On CTB, Bock hinge exceeds local CE by 0.86 percentage points in mean LAS and 1.52 points in mean LCM, and exceeds CL/E hinge by 0.78 and 2.01 points, respectively. These higher means accompany greater variation across seeds, with sample standard deviations of 1.08 points for LAS and 2.14 points for LCM. The paired LAS difference between Bock and CL/E hinge changes sign across seeds. These results describe the complete training and model-selection conditions; they establish no accuracy ordering between the inference algorithms on identical scores.
Earlier biaffine-parser experiments reported PTB LAS and LCM gains from CL/E hinge and decreases on CTB relative to their local baseline \citep{zhang-etal-2019-empirical}. Our CL/E-hinge condition has lower mean LCM than local CE on both corpora. The architectures and scoring protocols differ.

\section{Discussion and conclusion}
\label{sec:conclusion}

Bock's column adjustments, component labels, and star-and-bar transfers provide a matrix representation of exact arborescence inference. In exact arithmetic, the reformulated procedure reproduces the original procedure under matching scan and tie rules. The component-dependent cost relation explains the correspondence between candidate choices, while the recursive bar invariant establishes the equivalence of transfer and recovery. Under the prescribed component schedule and common original-arc tie order, Bock's starred arc set equals the branching returned by corrected Tarjan recovery at every completed column prefix. Their final rooted arborescences therefore coincide even when the optimum is nonunique. The separate analysis of Dozat-derived CL/E establishes optimal-value equivalence under the stated score and single-root conventions.

These results distinguish optimal values, returned trees, and learning trajectories. On identical scores and admissible trees, exact decoders attain the same optimum value; uniqueness additionally determines the returned tree. For structured-hinge training, \jp{on identical scores, gold trees, task costs, and admissible tree families, exact loss-augmented oracles define the same hinge objective}, and a unique maximizing tree determines the same score subgradient. Tied optima permit different valid subgradients. Identical training trajectories require agreement in the remaining computations and update decisions as well. Our training conditions use their corresponding decoders for development selection, so their accuracy differences compare complete training and model-selection procedures. 

For the local-CE models, the fixed-score comparison yields identical heads, full relation labels, and original tree-score values among CL/E, original Bock, and reformulated Bock on every evaluated development and test sentence. This empirical agreement concerns ordinary decoding from those models. The structured-training experiments use Bock and CL/E for loss-augmented inference, development selection, and final decoding on PTB and CTB. 
Our analysis complements algorithm-engineering comparisons of arborescence implementations, where practical performance depends on graph structure and data structures \citep{bother-kibig-weyand-2023-efficiently}. It establishes Bock as an alternative exact parsing and structured-training oracle, with a faithful reformulated implementation and explicit conditions governing its theoretical and empirical correspondence with CL/E.

\section*{Declaration of generative AI in the manuscript preparation process}

During the preparation of this work, the authors used OpenAI ChatGPT and Codex to assist with language editing, sentence restructuring, manuscript organization, and improving the clarity and consistency of the technical and mathematical exposition. The authors reviewed and edited all AI-assisted content, verified the formal arguments and implementations, and take full responsibility for the content of the article.



\appendix

\section{{Bock's original algorithm}}
\label{sec:bock-original}

Let \(V=\{1,\dots,N\}\) be a set of nodes with designated origin \(j_0\). Following Bock, we use \emph{link} for a directed arc and represent the directed network by an \(N\times N\) cost matrix \(C\). {For \(I,J\in V\), the entry \(C[I,J]\) is the cost of assigning head \(I\) to dependent \(J\), represented by the link \(I\to J\). Rows index heads and columns index dependents.} An entry \(C[I,J]=+\infty\) denotes a forbidden link. We write Bock's symbolic infinity as \(+\infty\), consistently with the main text; these entries are excluded from candidate selection and exceed every finite comparison value.

A feasible solution \(T\) contains exactly one link entering each node \(J\neq j_0\), no link entering \(j_0\), and a directed path from \(j_0\) to every other node. The minimum-cost arborescence problem is to find such a \(T\) minimizing \(\sum_{(I,J)\in T} C[I,J]\).

Bock processes the nonorigin columns in increasing index order. A link provisionally selected to enter node \(J\) is called \emph{starred}, and its {head} is recorded in \(I^{*}[J]\). When the iteration for column \(K\) terminates, every processed nonorigin column \(J\leq K\) has exactly one starred entering link, although circuit resolution may subsequently replace that link. Nodes carrying the same positive \(\mathrm{SPAN}\) label form a \emph{span}. Every node initially forms a singleton span; assigning a common label to several spans represents their contraction without changing the dimensions or indices of the cost matrix.

At the beginning of each candidate search, the starred links form a directed forest when every current span is treated as a single contracted node. {If the backward trace from a candidate's head reaches the active span, the starred links between spans form a directed path from the active span to the span containing that head. The candidate closes this path into a circuit of spans.} This invariant underlies Bock's circuit test and contraction operation. The principal arrays maintained by the procedure are summarized in Table~\ref{tab:bock-state}.

\begin{table*}[!ht]
\centering
\footnotesize
\begin{tabular}{r p{0.85\linewidth}@{}}
\toprule
Symbol & Role in Bock's procedure \\
\midrule
\(U_1[J]\) &
Amount subtracted from the costs of links entering \(J\). The {adjusted cost} of \(I\to J\) is \(C[I,J]-U_1[J]\); the procedure changes \(U_1[J]\) when the span containing \(J\) is adjusted. \\

\(I^{*}[J]\) &
{Head} of the link currently starred into \(J\). Thus, \(I^{*}[J]=I\) means that \(I\to J\) is provisionally selected, whereas \(I^{*}[J]=0\) means that no link is starred into \(J\). \\

\(\bar I[J],\bar J[J]\) &
Arc reference used by the star-and-bar transfer. {In a nonempty pair, \(\bar I[J]\) is the referenced link's head and \(\bar J[J]\) its dependent; \(\bar J[J]\) need not equal \(J\).} A backward trace writes its candidate into an empty entry, and L10 may subsequently redirect that reference to a displaced link. If a star entering \(J\) is displaced, the entry read at \(J\) supplies the next candidate. The pair \((0,0)\) denotes an empty entry. \\

\(\mathrm{SPAN}[J]\) &
Label of the span currently containing \(J\). Equal positive labels identify nodes represented as one contracted span; a negative label temporarily marks a span encountered during the backward trace. \\
\bottomrule
\end{tabular}
\caption{Principal arrays maintained by Bock's procedure.}
\label{tab:bock-state}
\end{table*}

The scalar \(K\) identifies the column currently being processed, and \(H=\mathrm{SPAN}[K]\) is the label of its active span; the columns \(J\leq K\) satisfying \(\mathrm{SPAN}[J]=H\) constitute that span in the current search. The scalar \(SS\) supplies a fresh label when spans are contracted. {The pair \((I_1,J_1)\) denotes the candidate link, with head \(I_1\) and dependent \(J_1\), and \(DU\) is its adjusted cost. At the beginning of each transfer pass, \((I_2,J_2)\) records the link displaced during the preceding pass.} Finally, \(Z\) stores the objective value of the completed arborescence. The starred array records the provisional link selection, \(U_1\) supplies the column adjustments used in the {adjusted-cost} comparisons, and the span and bar arrays implement circuit contraction and resolution.

{For the current column \(K\), the procedure chooses a link of minimum adjusted cost entering the active span from a different span, and adds that cost to every column adjustment in the active span.} It then begins at the {head} of the selected link and follows starred links backward. If the trace reaches the active span, the selected link closes a circuit with the starred path, so the active span and all spans encountered during the trace are contracted under a common label. Candidate selection is then repeated for the resulting span. If the trace instead reaches a node without a starred entering link, the candidate is installed. Any star thereby displaced is propagated through the candidate pairs stored in the bar arrays until a link is installed without displacing another star. The procedure then advances to the next nonorigin column.

Algorithm~\ref{alg:bock-overview} gives a line-mapped summary of this control flow. Labels of the form L2, L3, and so forth are the statement labels in Bock's original Algol procedure. The input and reporting statements at L1 and L97 are omitted.

\begin{algorithm*}[!ht]
\footnotesize
\caption{Control-flow summary of Bock's original procedure.}
\label{alg:bock-overview}
\begin{algorithmic}
\REQUIRE Number of nodes \(N\); origin \(j_0\); cost matrix \(C\) with \(+\infty\) for forbidden links
\ENSURE Starred array \(I^{*}\) defining a minimum-cost arborescence, or \textsc{Infeasible}

\STATE \textbf{L2 (initialization).}
For \(1\leq J\leq N\), set \(U_1[J]\), \(I^{*}[J]\), \(\bar I[J]\), and \(\bar J[J]\) to \(0\), and set \(\mathrm{SPAN}[J]\gets J\); set \(SS\gets N\), \(K\gets0\), and \(Z\gets0\).

\STATE \textbf{L3 (column advance).}
If \(K=N\), transfer control to L99; otherwise increment \(K\).
If \(K=j_0\), return to L3; otherwise continue to L4.

\STATE \textbf{L4 (candidate search).}
Set \(DU\gets +\infty\) and \(H\gets\mathrm{SPAN}[K]\). For every \(J\leq K\) with \(\mathrm{SPAN}[J]=H\) and every \(I\) with \(\mathrm{SPAN}[I]\neq H\) and \(C[I,J]<+\infty\), retain \((I_1,J_1)=(I,J)\) whenever \(C[I,J]-U_1[J]<DU\), and set \(DU\gets C[I,J]-U_1[J]\). If \(DU=+\infty\), transfer control to L98.

\STATE \textbf{L5 (column adjustment).}
For every \(J\leq K\) with \(\mathrm{SPAN}[J]=H\), set \(U_1[J]\gets U_1[J]+DU\).

\STATE \textbf{L6 (trace initialization).}
Set \(J\gets I_1\).

\STATE \textbf{L7 (circuit test and contraction).}
If \(\mathrm{SPAN}[J]=H\), \(SS\gets SS + 1\); for every \(q\leq K\) with \(\mathrm{SPAN}[q]=H\) or \(\mathrm{SPAN}[q]<0\), set \(\mathrm{SPAN}[q]\gets SS\); then return to L4.

\STATE \textbf{L8 (trace continuation or exit).}
{If \(I^{*}[J]>0\), perform the following operations in order: (i) if \(\mathrm{SPAN}[J]>0\), set \(H_1\gets\mathrm{SPAN}[J]\) and negate \(\mathrm{SPAN}[q]\) for every \(q\leq K\) satisfying \(\mathrm{SPAN}[q]=H_1\); (ii) if \(\bar I[J]=0\), set \((\bar I[J],\bar J[J])\gets(I_1,J_1)\); and (iii) set \(J\gets I^{*}[J]\) and return to L7. If \(I^{*}[J]=0\), fall through to L9.}

\STATE \textbf{L9 (transfer initialization).}
Restore all negative span labels and set \((I_2,J_2)\gets(0,0)\).

\STATE \textbf{L10--L11 (star-and-bar transfer).}
{For every \(q\leq K\) satisfying \((\bar I[q],\bar J[q])=(I_1,J_1)\), set \((\bar I[q],\bar J[q])\gets(I_2,J_2)\). Set \((I,J)\gets(\bar I[J_1],\bar J[J_1])\), then set \((\bar I[J_1],\bar J[J_1])\gets(I_2,J_2)\). Set \(I_2\gets I^{*}[J_1]\) and \(I^{*}[J_1]\gets I_1\). If \(I_2=0\), return to L3; otherwise set \((J_2,I_1,J_1)\gets(J_1,I,J)\) and return to L10.}

\STATE \textbf{L98 (infeasibility).}
Report \textsc{Infeasible}, set \(Z\gets +\infty\), and transfer control to L100.

\STATE \textbf{L99 (objective value).}
Set \(Z\gets\sum_{J\neq j_0}C[I^{*}[J],J]\) and report an optimum solution.

\STATE \textbf{L100 (output).}
Output \(Z\), \(U_1\), \(I^{*}\), \(\bar I\), \(\bar J\), and \(\mathrm{SPAN}\).
\end{algorithmic}
\end{algorithm*}

\paragraph{Candidate selection and column adjustment}
At L4, the active-span label is \(H=\mathrm{SPAN}[K]\). The procedure examines columns \(J\leq K\) satisfying \(\mathrm{SPAN}[J]=H\) and feasible links \(I\to J\) satisfying \(\mathrm{SPAN}[I]\neq H\). Among these links, it selects \((I_1,J_1)\) with least {adjusted cost} \(C[I,J]-U_1[J]\). Because the candidate is replaced only under a strict inequality, ties are resolved by the first minimum encountered in Bock's scan order: {dependents in increasing \(J\), with heads in increasing \(I\) for each dependent.} If no candidate is found, L98 reports the instance infeasible. Otherwise, L5 adds \(DU\) to the column adjustment of every column in the active span. Immediately after this update, the selected link satisfies \(C[I_1,J_1]-U_1[J_1]=0\) in the current search state. Later adjustments to a contracted span may alter the {adjusted cost} of a previously starred link.

\paragraph{Backward trace and contraction}
{The original variable \(J\) is reused as the current node during backward tracing. At L6 it is initialized to the candidate's head \(I_1\), while \(J_1\) remains the candidate's dependent.} If \(\mathrm{SPAN}[J]=H\), the backward trace has returned to the active span. {By the starred-forest invariant, the traced links between spans form a directed path from the active span to the span containing \(I_1\). The candidate \(I_1\to J_1\), whose dependent \(J_1\) lies in the active span, closes this path into a circuit of spans.} At L7, the procedure increments \(SS\) and assigns the new label to the active span and every span marked during the trace. It then returns to L4 to search for a link entering the resulting contracted span.

If \(\mathrm{SPAN}[J]\neq H\) and \(I^{*}[J]>0\), L8 continues the trace. If \(\mathrm{SPAN}[J]>0\), the span has not yet been marked: L8 sets \(H_1\gets\mathrm{SPAN}[J]\) and negates \(\mathrm{SPAN}[q]\) for every \(q\leq K\) whose span label is \(H_1\). If the bar entry at \(J\) is empty, L8 records the current candidate by setting \((\bar I[J],\bar J[J])\gets(I_1,J_1)\). {A nonempty entry is left unchanged. A newly written entry names the candidate whose backward trace has reached \(J\). Contraction preserves the reference, while L10 may subsequently redirect it before it is read as a transfer candidate.} The trace then follows the starred link backward by setting \(J\gets I^{*}[J]\) and returning to L7. Repeating L7--L8 records the candidate at each encountered node whose bar entry is empty. If \(I^{*}[J]=0\), the trace has reached a root of the current starred forest, and control falls through to L9.

\paragraph{Star-and-bar transfer}
The transfer is an exchange chain. {At the beginning of each pass, \((I_1,J_1)\) identifies the candidate link and \((I_2,J_2)\) the link displaced during the preceding pass; the latter is initialized to \((0,0)\) at L9.} If starring \(I_1\to J_1\) displaces the existing star into \(J_1\), the bar entry at \(J_1\) supplies the next candidate in the chain.

Each pass through L10 first replaces every bar entry equal to the current candidate \((I_1,J_1)\) by the previously displaced pair \((I_2,J_2)\), thereby redirecting references from the link being installed to the link displaced during the preceding pass. It then reads the candidate stored at \(J_1\) into \((I,J)\), replaces that bar entry by \((I_2,J_2)\), saves the {head of the existing starred link} as \(I_2\gets I^{*}[J_1]\), and installs the current candidate by setting \(I^{*}[J_1]\gets I_1\).

If \(I_2=0\), the installation has not displaced a star, so the exchange chain terminates and control returns to L3. Otherwise, the displaced link is \((I_2,J_1)\), the stored pair \((I,J)\) becomes the next candidate, and the assignment \((J_2,I_1,J_1)\gets(J_1,I,J)\) prepares the next pass through L10. The transfer therefore follows the bar references {maintained by backward tracing and global replacement} until a candidate is installed at a node without a starred entering link.

When L3 finds \(K=N\), L99 computes \(Z=\sum_{J\neq j_0}C[I^{*}[J],J]\). On feasible termination, \(I^{*}[J]=I\) specifies the selected link \(I\to J\) for every \(J\neq j_0\), while \(I^{*}[j_0]=0\).

\paragraph{Example}
Consider \(N=3\), origin \(j_0=1\), and
\(C=\left(\begin{smallmatrix}M&50&60\\M&M&10\\M&20&M\end{smallmatrix}\right)\).
The feasible links are \(1\to2\) of cost \(50\), \(1\to3\) of cost \(60\), \(2\to3\) of cost \(10\), and \(3\to2\) of cost \(20\). Although small, this instance exercises both circuit contraction and the subsequent transfer. After L2, \(U_1=(0,0,0)\), \(I^{*}=(0,0,0)\), \(\bar I=\bar J=(0,0,0)\), and \(\mathrm{SPAN}=(1,2,3)\); L3 skips the origin column.

For \(K=2\), the active span is \(H=2\). L4 selects \(3\to2\) with \(DU=20\), and L5 sets \(U_1[2]=20\). The trace begins at node \(3\); because \(I^{*}[3]=0\), it reaches L9, and L10 stars \(3\to2\), giving \(I^{*}[2]=3\). Because the trace encounters no starred link, L8 records no bar pair, and \(\bar I=\bar J=(0,0,0)\).

For \(K=3\), the active span is \(H=3\). L4 selects \(2\to3\) with \(DU=10\), and L5 sets \(U_1[3]=10\). {The trace begins at \(J=2\). Since \(\mathrm{SPAN}[2]=2\neq H\) and \(I^{*}[2]=3>0\), L8 changes \(\mathrm{SPAN}[2]\) from \(2\) to \(-2\), records \((\bar I[2],\bar J[2])=(2,3)\), and sets \(J\gets3\). On returning to L7, \(\mathrm{SPAN}[3]=H\); the trace has therefore closed the circuit formed by \(3\to2\) and \(2\to3\). L7 assigns the new label \(4\) by \(SS + 1\) to the active span and the negatively marked span, giving \(\mathrm{SPAN}=(1,4,4)\), while leaving the bar entry at node \(2\) unchanged.}

The renewed search compares links entering the contracted span from nodes outside it; in this example, the only such node is \(1\).
Their reduced costs are \(50-U_1[2]=30\) for \(1\to2\) and \(60-U_1[3]=50\) for \(1\to3\). Hence, \(1\to2\) is selected with \(DU=30\), and L5 yields \(U_1=(0,50,40)\). {The trace begins at \(J=1\); because \(I^{*}[1]=0\), control reaches L9, which initializes \((I_2,J_2)=(0,0)\). On the first pass through L10, the procedure reads \((\bar I[2],\bar J[2])=(2,3)\), clears this entry, displaces \(3\to2\), and stars \(1\to2\). The displaced pair becomes \((I_2,J_2)=(3,2)\), and \(2\to3\) becomes the next candidate. On the second pass, L10 stores \((3,2)\) in the empty bar entry at node \(3\) and stars \(2\to3\). Since \(I^{*}[3]\) was \(0\), no further star is displaced and the transfer terminates.}

The final starred array \(I^{*}=(0,1,2)\) records no entering link for origin \(1\) and selects \(1\to2\) and \(2\to3\). The remaining bar entry is residual bookkeeping and does not affect this arborescence. L99 therefore obtains \(Z=50+10=60\).

\section{Experimental details}
\label{sec:experimental-details}

\paragraph{Corpora and model}

We use dependency conversions of the English Penn Treebank and Chinese Treebank 5.1, supplied in CoNLL-U format. PTB uses WSJ sections 02--21 for training, section 22 for development, and section 23 for testing, containing 39,832, 1,700, and 2,416 sentences, respectively. CTB follows the standard dependency-parsing split \citep{zhang-clark:2008:EMNLP}: training documents 001--815 and 1001--1136, development documents 886--931 and 1148--1151, and test documents 816--885 and 1137--1147. We exclude four CTB training sentences containing 206 tokens because of malformed annotations, retaining 16,087 training sentences and 437,778 tokens. The 803 development and 1,910 test sentences are unchanged. Every condition uses the same corpus files.

The shared Stanza parser has a three-layer bidirectional LSTM with hidden-size setting 400 \citep{dozat-manning-2017-deep,qi-etal-2020-stanza}. Word and character representations, UPOS, XPOS, and morphological features are identical across conditions. We use gold tokenization and the supplied linguistic features. External static word embeddings and character language models are disabled. BERT is fully fine-tuned in every run \citep{devlin-etal-2019-bert}, using \texttt{bert-base-cased} for PTB and \texttt{bert-base-chinese} for CTB.

\paragraph{Training and model selection}

Each condition starts independently from the same pretrained BERT model for its language, using seeds 1111, 2222, and 3333. The three seeds run concurrently on a single physical GPU, an NVIDIA A100 80GB PCIe, with conditions executed sequentially within each seed. Deterministic algorithms are disabled. Shared seeds and pretrained-model identifiers do not establish identical initialized tensors or numerical execution.

Training uses a token budget of 5,000 per batch, at most 50,000 updates, 500 BERT warm-up steps, and Stanza's Adadelta-to-Adam schedule. The hinge margin is one per incorrect head. Relation-label cross-entropy, auxiliary losses, and loss normalization remain unchanged when the arc cross-entropy is replaced by the hinge.

Development evaluation occurs every 100 updates. \jp{For the conditions in the main results table, local CE and CL/E hinge use CL/E for development and final decoding; Bock hinge uses reformulated Bock for both. The additional local-CE runs use their designated decoder---CL/E, original Bock, or reformulated Bock---for development and final decoding, with no structured-training oracle.} We retain the model with the highest development LAS, preferring the later model when scores tie. Development scores also govern learning-rate scheduling, optimizer transitions, and stopping. \jp{At the transition to Adam, training reloads the development-selected best model.}

\jp{The first recorded training loss is at update 20, before development evaluation at update 100. Each recorded value is the current minibatch's total normalized loss, including arc, relation, and auxiliary terms, rather than an isolated hinge value.}

\paragraph{Structured-hinge training}
\label{sec:structured-training}

{Let $T^\star\in\mathcal T_1$ be the gold tree and write $S(T)=\sum_{(i,j)\in T}S[i,j]$. The margin-rescaled structured hinge is \citep{tsochantaridis-EtAl:2004:ICML,taskar-etal-2004-max}}
\begin{equation}
\begin{aligned}
  L(S,T^\star)
  &=\max_{T\in\mathcal T_1}
    \bigl[S(T)+\Delta(T,T^\star)\bigr]\\
  &\quad-S(T^\star).
\end{aligned}
\label{eq:structured-hinge}
\end{equation}
{Here $\Delta$ is $\mu\geq0$ times the number of incorrect heads. Gold-tree feasibility gives $L\geq0$. Loss-augmented inference uses}
\begin{equation}
  \widetilde S[i,j]
  =S[i,j]+\mu\,\mathbf1[(i,j)\notin T^\star]
  \label{eq:loss-augmented-scores}
\end{equation}
on admissible arcs; forbidden arcs remain excluded. Before inference, column-wise log-softmax is applied to the augmented matrix. It subtracts a constant from each dependent column and therefore preserves all maximizing trees \jp{in exact arithmetic. Both oracle implementations use this normalization; identical normalized inputs are assumed in the equivalence theorem below.}

{The maximizing tree $\widehat T$ is selected outside automatic differentiation. Evaluating the loss from the original neural scores while holding $\widehat T$ fixed gives the score subgradient}
\begin{equation}
\begin{aligned}
  g[i,j]
  &=\mathbf1[(i,j)\in\widehat T]\\
  &\quad-\mathbf1[(i,j)\in T^\star]
\end{aligned}
\label{eq:hinge-subgradient}
\end{equation}
{on admissible score coordinates.}

\begin{theorem}[Structured-hinge equivalence]
\label{thm:hinge-equivalence}
{Suppose both oracles receive identical scores, gold tree, margin, and admissible arcs, and the common normalized loss-augmented matrix satisfies the hypotheses of Theorem~\ref{thm:single-root-equivalence}. In exact arithmetic, Bock and CL/E compute the same value of \eqref{eq:structured-hinge}. Every selected maximizer gives a valid score subgradient, and a unique loss-augmented optimum forces the same selected score subgradient.}
\end{theorem}

\begin{proof}
{Normalization preserves the maximizing trees, and Theorem~\ref{thm:single-root-equivalence} makes both oracles exact over the same tree family. They therefore attain the same loss-augmented optimum. The affine objective associated with a selected maximizer equals the hinge at the current scores and lies below it everywhere, so its slope \eqref{eq:hinge-subgradient} is a subgradient. Distinct maximizing trees have distinct arc indicators, while uniqueness forces the same choice.}
\end{proof}

{Matching initial model and optimizer states, minibatches, stochastic and numerical computations, auxiliary losses, selected score subgradients, update rules, and development-dependent decisions yields identical training trajectories: equal states give equal forward computations, parameter gradients, and updates, so induction preserves equality.}

\paragraph{Evaluation}

We compute UAS and LAS with the bundled CoNLL 2018 evaluator \citep{zeman-etal-2018-conll}. Scoring uses integer-ID tokens, includes punctuation, and excludes the artificial ROOT. LAS ignores dependency-relation subtypes. LCM applies the same label convention and counts a sentence as correct exactly when every scored token has both the correct head and the correct base dependency relation.

We reevaluate the saved test predictions without retraining, verify their token alignment with the gold files, and check attachment counts against the evaluator. Per-seed rates are computed from integer counts. Means and sample standard deviations are calculated from full-precision rates and rounded only for presentation.

For CTB, the reformulated-Bock-hinge minus CL/E-hinge LAS differences are $+1.32$, $-0.22$, and $+1.25$ percentage points for seeds 1111, 2222, and 3333, respectively. The higher aggregate mean therefore includes one paired run with lower LAS.

\paragraph{Comparison with earlier structured-hinge experiments}
\label{sec:earlier-hinge-comparison}

Earlier biaffine-parser experiments compare Local-Prob with Global-NProj-Hinge, using CL/E for nonprojective structured inference \citep{zhang-etal-2019-empirical}. On PTB, their LAS/LCM scores are 93.75/46.65 for Local-Prob and 93.91/47.78 for Global-NProj-Hinge. On CTB, the corresponding scores are 88.16/33.16 and 87.92/32.77.

\jp{The earlier models use fastText embeddings and character CNNs, combine arc and label scores, and use predicted PTB POS tags \citep{zhang-etal-2019-empirical}. They exclude punctuation in PTB and CTB evaluation, whereas our main results include it.}

We additionally reevaluate our saved predictions using the earlier POS-based punctuation exclusions. For PTB, excluded gold XPOS tags are \texttt{``}, \texttt{''}, \texttt{:}, \texttt{,}, and \texttt{.}; for CTB, the excluded tag is \texttt{PU}. All test sentences are retained.

Under these exclusions, PTB local CE obtains 95.10 LAS and 57.71 LCM, compared with 95.02 LAS and 56.91 LCM for CL/E hinge. CTB local CE obtains 90.56 LAS and 44.24 LCM, compared with 90.61 LAS and 43.73 LCM for CL/E hinge. \jp{Punctuation exclusion preserves the directions of these within-study differences. These are alternative evaluations of the same saved predictions.}

\end{document}